\documentclass[journal]{IEEEtran}

\usepackage{amsfonts}
\usepackage{amssymb}
\usepackage{amsmath}
\usepackage[utf8]{inputenc}
\usepackage[english]{babel}
\usepackage{algorithm}
 \usepackage{algorithmic}
\usepackage[FIGTOPCAP]{subfigure}
\addto\captionsenglish{%
  }
\DeclareMathOperator{\diag}{diag}

\usepackage{xspace}
\usepackage{intmacros}

\usepackage[final]{graphicx}

\newcommand{\lone}{$\ell^{1}$}
\newcommand{\ltwo}{$\ell^{2}$}

\newcounter{examplecounter}
\renewcommand{\theexamplecounter}{\arabic{examplecounter}}

\makeatletter
\@ifclasslater{IEEEtran}{2012/12/26}
 {\newcommand{\killproofname}{\unskip\nopunct}}
 {\newcommand{\killproofname}[1]{\unskip\aftergroup\ignorespaces\ignorespaces}}
\makeatother

\ifCLASSINFOpdf
\else
\fi

\begin{document}
%
\title{Bayesian Hypothesis Testing for Block Sparse Signal Recovery}
%
%
%

\author{Mehdi~Korki,~\IEEEmembership{Student Member,~IEEE,}
Hadi~Zayyani, and~Jingxin~Zhang
\thanks{M. Korki and J. Zhang are with the Department
of Telecommunications, Electrical, Robotics and Biomedical Engineering, Swinburne University of Technology, Hawthorn,
3122 Australia (e-mail: mkorki@swin.edu.au; jingxinzhang@swin.edu.au).}
\thanks{H. Zayyani is with the Department of Electrical and Computer Engineering, Qom University of Technology, Qom, Iran  (e-mail: zayyani2009@gmail.com).}} 
\maketitle

\begin{abstract}
This letter presents a novel Block Bayesian Hypothesis Testing Algorithm (Block-BHTA) for reconstructing block-sparse signals with unknown block structures. The Block-BHTA comprises 
the detection and recovery of the supports, and the estimation of the amplitudes of the block sparse signal. The support detection and recovery is performed using a Bayesian hypothesis
testing. Then, based on the detected and reconstructed supports, the nonzero amplitudes are estimated by linear MMSE. The effectiveness of Block-BHTA is demonstrated by numerical 
experiments.
\end{abstract}


%
\IEEEpeerreviewmaketitle

\section{Introduction}
%
%
%
%
\IEEEPARstart{C}ompressed sensing (CS) and sparse signal recovery aim to recover the sparse signal, a signal with only a few nonzero elements,
from underdetermined systems of linear equations. In some applications,  the unknown signal to be estimated has additional structure.
If the structure of the signal is exploited, the better recovery performance can be achieved. A block-sparse signal,
in which the nonzero samples manifest themselves as clusters, is an important structured
sparsity \cite{refe_BHT_1}\nocite{refe_BHT_2}\nocite{refe_7}--\cite{refe_8}. Block-sparsity has a wide range of applications in multiband signals~\cite{refe_1},
audio signals~\cite{refe_2}, structured
compressed sensing~\cite{refe_3}, and the multiple measurement vector (MMV) model~\cite{refe_4}. The general mathematical model of the block sparse signal is
\begin{equation}
\label{equ_1}
\mathbf{y}=\mathbf{\Phi}\mathbf{w}+\mathbf{n},
\end{equation}
where $\mathbf{\Phi} \in \mathbb{R}^{N\times M}$ is a known measurement matrix,
$\mathbf{y} \in \mathbb{R}^{N}$ is the available measurement vector, and $\mathbf{n} \in \mathbb{R}^{N}$ is
the Gaussian corrupting noise. We aim to  estimate the original unknown signal $\mathbf{w} \in \mathbb{R}^{M}$, when $N\ll M$, with the cluster structure
\begin{equation}
\label{equ_2}
\mathbf{w}=[\underbrace{w_{1},\ldots,w_{d_{1}}}_{\mathbf{w}^{T}[1]},\ldots,\underbrace{w_{d_{g-1}+1},
\ldots,w_{d_{g}}}_{\mathbf{w}^{T}[g]}]^{T},
\end{equation}
where $\mathbf{w}[i]$ denotes the $i$th block with length $d_{i}$ which are not necessarily identical. In the
block partition (\ref{equ_2}), only $k\ll g$ vectors $\mathbf{w}[i]$ have nonzero Euclidean norm.

Given the {\it a priori}
knowledge of
block partition, a few algorithms such as  Block-OMP \cite{refe_BHT_1}, mixed \ltwo/\lone norm-minimization \cite{refe_BHT_2}, group LASSO \cite{refe_7}
and model-based CoSaMP \cite{refe_8}, work effectively in the block-sparse signal recovery. 
These algorithms require the knowledge of
the block structure (e.g. the location and the lengths of the blocks) in (\ref{equ_2}). However, in many applications, such prior knowledge is often unavailable. 
Hence, devising an adaptive method for estimating the block partition and recovering the clustered-sparse signal simultaneously remains a challenge.
To recover the structure-agnostic block-sparse signal, some algorithms, e.g. CluSS-MCMC \cite{refe_9}, BM-MAP-OMP \cite{refe_10}, Block Sparse Bayesian Learning (BSBL) \cite{refe_11}, and pattern-coupled SBL (PC-SBL) \cite{refe_BHT_7} have been proposed recently, which require less {\it a priori} information.

In this letter, we propose a novel Block Bayesian Hypothesis Testing algorithm (Block-BHTA) which uses a joint detection of the supports and estimation of the amplitudes. Block-BHTA utilizes a Bayesian hypothesis testing (BHT) for the detection and recovery of the supports. BHT was first proposed by Zayyani {\it et. al.} \cite{refe_BHT_3} in a Bayesian pursuit algorithm (BPA) for sparse representations. Recently, BHT with belief propagation has been introduced in noisy sparse recovery \cite{refe_BHT_4}. 

Inspired by BPA \cite{refe_BHT_3}, we adopt a BHT-based approach
and extend BPA to the block sparse recovery case (Block-BHTA). BPA uses the correlations between measurement vector $\mathbf{y}$ and the columns of matrix $\mathbf{\Phi}$ and applies a 
binary BHT to obtain an activity rule in which the correlations are compared with a threshold. This activity rule is then used for the detection and recovery of the supports.
Different to BPA, Block-BHTA searches for the start and termination of the blocks of the supports in the block-sparse signal $\mathbf{w}$. This search, performed by the BHT, leads to two ultimate 
activity rules where the correlations between measurement vector $\mathbf{y}$ and the columns of matrix $\mathbf{\Phi}$ manifest themselves in these two activity rules. Hence, the
correlations play an important role in both BPA and Block-BHTA. In these two activity rules, the correlations are compared with two simple thresholds to detect and recover the supports. Given the detected and recovered supports, Block-BHTA then uses a linear MMSE to estimate the nonzero amplitudes. Block-BHTA also uses Bernoulli-Gaussian hidden Markov model (BGHMM) \cite{refe_40} for the block-sparse signals. Using simple tuning updates, Block-BHTA utilizes a maximum {\it a posteriori} (MAP) estimation procedure to automatically learn all parameters of the statistical signal model (e.g. the variance and the elements of state-transition matrix of BGHMM). The efficiency of the proposed Block-BHTA is verified by numerical experiments.

The rest of the letter is organized as follows. In Section \ref{sec_sigmodel}, we present the signal model. In Section \ref{sec_opt}, the Block-BHTA  is
proposed. Experimental results are presented in Section \ref{sec_sim}. Finally, conclusions are drawn in Section \ref{sec_conc}.
\section{Signal Model}\label{sec_sigmodel}

Consider the linear model of (\ref{equ_1}) as the measurement process of an underlying time- or spatial-series which is non-i.i.d and block sparse. The measurement matrix $\mathbf{\Phi}$ is assumed known and its
columns are normalized to have unit norms. Furthermore, we model the noise in (\ref{equ_1}) as a stationary, additive white Gaussian noise (AWGN) process, with
$\mathbf{n}\sim \mathcal{N}\left(0,\sigma^{2}_{n}\boldsymbol{I}_{N}\right)$. To model the block-sparse sources ($\mathbf{w}$), we introduce two hidden random processes, $\mathbf{s}$
and
$\boldsymbol{\theta}$ \cite{refe_41}, \cite{refe_21}. The binary vector $\mathbf{s} \in \left\{0,1\right\}^{M}$ describes the support of $\mathbf{w}$, denoted $\mathcal{S}$, while the vector $\boldsymbol{\theta} \in
\mathbb{R}^{M}$ represents the amplitudes of the active elements of $\mathbf{w}$. Hence, each element of the source vector $\mathbf{w}$ can be characterized as
\begin{equation}
\label{equ_3}
w_{i}=s_{i}\cdot \theta_{i},
\end{equation}
where $s_{i}=0$ gives $w_{i}=0$ for $i \notin \mathcal{S}$ and $s_{i}=1$ gives $w_{i}=\theta_{i}$ for $i \in \mathcal{S}$. In vector form, (\ref{equ_3}) can be written as
$\label{equ_vec} \mathbf{w}=\mathbf{S}\boldsymbol{\theta}$, where $\mathbf{S}=\diag(\mathbf{s}) \in\mathbb{R}^{M \times M}$.

To model the block-sparsity of the source vector $\mathbf{w}$, we assume that its supports are correlated such that $\mathbf{s}$ is a stationary first-order Markov process defined by two transition probabilities:
$p_{10}\triangleq \mathrm{Pr}\left\{s_{i+1}=1|s_{i}=0\right\}$ and $p_{01}\triangleq \mathrm{Pr}\left\{s_{i+1}=0|s_{i}=1\right\}$. Therefore, in the steady state, $\mathrm{Pr}\left\{s_{i}=0\right\}= p=\frac{p_{01}}{p_{10}+p_{01}}$ and $\mathrm{Pr}\left\{s_{i}=1\right\}=1- p=\frac{p_{10}}{p_{10}+p_{01}}$, which determine the probabilities of the states in relation to the transition probabilities. The two parameters $p$ and $p_{10}$ completely describe the state process of the Markov chain. As a result, the remaining transition probability can be determined as
$p_{01}=\frac{p\cdot p_{10}}{(1-p)}$.
The length of the blocks of the block-sparse signal is determined by parameter $p_{01}$, namely, the average number of consecutive samples of ones is specified by $1/p_{01}$ in the Markov chain.

We further assume that the amplitude vector $\boldsymbol{\theta}$ has a Gaussian distribution with $\boldsymbol{\theta}\sim \mathcal{N}\left(0,\sigma^{2}_{\theta}\boldsymbol{I}_{M}\right)$.
Hence, the PDF of the $w_i$'s is given as
\begin{equation}
\label{equ_4}
p(w_{i})=p\delta(w_{i})+(1-p)\mathcal{N}\left(w_{i};0,\sigma^{2}_{\theta}\right),
\end{equation}
where $\sigma^{2}_{\theta}$ is the variance of $\boldsymbol{\theta}$.

Equation (\ref{equ_4}) is the well known BGHMM which is a special form of Gaussian Mixture Hidden Markov model (GHMM). The hidden variables $s_{i}$ with the first-order Markov chain model in BGHMM allow implicit expression of the block-sparsity of the signal $\mathbf{w}$ to be estimated. 

\section{The Proposed Algorithm}\label{sec_opt}

The proposed Block-BHTA consists of support detection and amplitude estimation. Using BHT, we first detect and recover the Block-sparse support $\mathbf{s}$. Then, using a linear MMSE estimator, we estimate the non-zero amplitudes of the detected supports (i.e., estimating $\boldsymbol{\theta}$). 

\subsection{Support Detection Using Bayesian Hypothesis Testing}\label{sec_supp}
We determine the activity of the $j$th element of the block-sparse signal $\mathbf{w}$ by searching the start and termination of active blocks in $\mathbf{w}$. Toward that end,
we assume that $w_{i}$ is inactive (i.e., $s_{i}=0$) and we intend to determine whether $w_{i+1}$ is active (i.e., $s_{i+1}=1$). This case is equivalent to searching the start of the
active blocks. In the second case, we assume that $w_{i}$ is active (i.e., $s_{i}=1$) and we intend to determine whether $w_{i+1}$ is inactive (i.e., $s_{i+1}=0$). This
corresponds to searching the end of active blocks. Full details are given below.

\subsubsection{Searching The Start of Active Blocks}
In order to detect the start of an active block we choose one between the hypotheses $\mathcal{H}_{01}: s_{i}=0,s_{i+1}=1$ and $\mathcal{H}_{00}: s_{i}=0,s_{i+1}=0$, given the 
measurement vector $\mathbf{y}$. The Bayesian hypothesis test is
\begin{equation}
\label{equ_5}
\widehat{s}_{j}=
\begin{cases}
1 &  p\left ( \mathcal{H}_{01,j}\mid \mathbf{y} \right )>p\left ( \mathcal{H}_{00,j}\mid \mathbf{y} \right ), \\
0 &  Otherwise,
\end{cases}
\end{equation}
where $\mathbf{y}$ is the measurement vector. The posterior probability $p\left ( \mathcal{H}_{01,j}\mid \mathbf{y} \right )$ is given as
\begin{align}
\label{equ_6}
p\left ( \mathcal{H}_{01,j}\mid \mathbf{y} \right )&=p\left ( s_{i}s_{i+1}=01\mid \mathbf{y} \right)=p\left ( s_{i}=0 \right )\nonumber\\
&\quad\times p\left ( s_{i+1}=1 \mid s_{i}=0 \right ) \times p\left ( \mathbf{y}\mid s_{i}s_{i+1}=01\right )\nonumber\\
&=p \times p_{10} \times p\left ( \mathbf{y}\mid s_{i}s_{i+1}=01\right ),
\end{align}
where $\mathbf{y}=\sum_{j=1,j\neq i }^{M}\boldsymbol{\phi}_{j}w_{j}+\mathbf{n}$ and $\boldsymbol{\phi}_{j}$ represents the $j$th column of matrix $\mathbf{\Phi}$. Similarly, the posterior probability
$p\left ( \mathcal{H}_{00,j}\mid \mathbf{y} \right )$ is given by
\begin{align}
\label{equ_7}
p\left ( \mathcal{H}_{00,j}\mid \mathbf{y} \right )&=p\left ( s_{i}s_{i+1}=00\mid \mathbf{y} \right)=p\left ( s_{i}=0 \right )\nonumber\\
&\quad\times p\left ( s_{i+1}=0 \mid s_{i}=0 \right ) \times p\left ( \mathbf{y}\mid s_{i}s_{i+1}=00\right )\nonumber\\
&=p \times p_{00} \times p\left ( \mathbf{y}\mid s_{i}s_{i+1}=00\right ),
\end{align}
where $p_{00}=p\left ( s_{i+1}=0 \mid s_{i}=0 \right )=1-p_{10}$ and $\mathbf{y}=\sum_{j=1,j\neq i ,i+1}^{M}\boldsymbol{\phi}_{j}w_{j}+\mathbf{n}$. Hence, from 
(\ref{equ_5})-(\ref{equ_7}), the activity rule for $w_{i+1}$ is 
\begin{equation}
\label{equ_8}
p_{10} \times p\left ( \mathbf{y}\mid s_{i}s_{i+1}=01\right )>p_{00} \times p\left ( \mathbf{y}\mid s_{i}s_{i+1}=00\right ).
\end{equation}
Assume that we have all the estimates of $w_{j}$ except for $j\neq i+1$ and we intend to estimate $w_{i+1}$. We have
\begin{equation}
\label{equ_9}
\begin{split}
p\left (\mathbf{y}| s_{i}s_{i+1} =00  \right )&=\frac{\exp\left (-\frac{1}{2\sigma_{n}^{2} }\left \|\mathbf{y}-\sum_{j=1,j\neq i,i+1 }^{M}\boldsymbol{\phi}_{j} w_{j}  \right \|_{2}^{2}  \right )}{\sqrt{\left (2\pi\sigma_{n}^{2}   \right )^{N}}}.
\end{split}
\end{equation}
When $s_{i}s_{i+1}=01$, we have $\mathbf{y}=\sum_{j=1,j\neq i,i+1}^{M}\boldsymbol{\phi}_{j}w_{j}+\boldsymbol{\phi}_{i+1}w_{i+1}+\mathbf{n}=\sum_{j=1,j\neq i,i+1}^{M}\boldsymbol{\phi}_{j}w_{j}+{\mathbf{n}}'$, where ${\mathbf{n}}'=\boldsymbol{\phi}_{i+1}w_{i+1}+\mathbf{n}$. Hence, the likelihood $p\left ( \mathbf{y}\mid s_{i}s_{i+1}=01\right )$ is a multivariate Gaussian with its mean and covariance given respectively by
\begin{equation}
\label{equ_10}
\boldsymbol{\mu} _{\mathbf{y}}=\sum_{j=1,j\neq i,i+1}^{M}\boldsymbol{\phi}_{j}w_{j},
\end{equation}
\begin{equation}
\label{equ_11}
\boldsymbol{\Sigma }_{\mathbf{y}}=\mathrm{Cov}\left ({\mathbf{n}}' \right )=\sigma^{2}_{n}\boldsymbol{I}_{N}+\sigma^{2}_{\theta}\boldsymbol{\phi}_{i+1}\boldsymbol{\phi}^{T}_{i+1}.
\end{equation}
Therefore, we can write the likelihood function as 
\begin{equation}
\label{equ_12}
p\left ( \mathbf{y}\mid s_{i}s_{i+1}=01\right )=\frac{\exp\left(-\frac{1}{2}\left ( \mathbf{y}- \boldsymbol{\mu} _{\mathbf{y}}\right )^{T}{\boldsymbol{\Sigma}^{-1}_{\mathbf{y}}}\left ( \mathbf{y}- \boldsymbol{\mu} _{\mathbf{y}}\right )\right)}{\sqrt{\left ( 2\pi \right )^{N}\det(\boldsymbol{\Sigma}_{\mathbf{y}})}}.
\end{equation}
Using the matrix inversion lemma (\cite{refe_BHT_6}, p. 571), we can express ${\boldsymbol{\Sigma}^{-1}_{\mathbf{y}}}$ as
\begin{equation}
\label{equ_13}
{\boldsymbol{\Sigma}^{-1}_{\mathbf{y}}}=\sigma^{-2}_{n}\boldsymbol{I}_{N}-\frac{\boldsymbol{\phi}_{i+1}\boldsymbol{\phi}^{T}_{i+1}\sigma^{-2}_{n}}{1+\left ( \frac{\sigma_{n} }{\sigma_{\theta } } \right )^{2}}.
\end{equation}
The determinant of ${\boldsymbol{\Sigma}_{\mathbf{y}}}$ can be calculated as \footnote{We have used matrix determinant lemma, i.e. $\det\left ( \mathbf{A}+\mathbf{u}\mathbf{v}^{T} \right )=\left ( 1+\mathbf{v}^{T}\mathbf{A}^{-1} \mathbf{u}\right )\det\left ( \mathbf{A}\right )$, where $\mathbf{A}$ is an invertible square matrix and $\mathbf{u}$, $\mathbf{v}$ are column vectors.}
\begin{align}
\label{equ_14}
\det\left ( {\boldsymbol{\Sigma}_{\mathbf{y}}} \right )&=
\left (  \sigma_{\theta }\right )^{2N} \det\left ( \frac{\sigma_{n}^{2} }{\sigma_{\theta }^{2} }\boldsymbol{I}_{N}+\boldsymbol{\phi }_{i+1}\boldsymbol{\phi }_{i+1}^{T} \right )\nonumber\\
&=\left (  \sigma_{\theta }\right )^{2N}\left ( 1+ \boldsymbol{\phi }_{i+1}^{T} \frac{\sigma_{n}^{2} }{\sigma_{\theta }^{2} }\boldsymbol{\phi }_{i+1}\right )\det\left ( \frac{\sigma_{n}^{2} }{\sigma_{\theta }^{2} }\boldsymbol{I}_{N} \right )\nonumber\\
&=\left (  \sigma_{\theta }\right )^{2N}\left ( 1+\frac{\sigma_{\theta}^{2} }{\sigma_{n}^{2} }\boldsymbol{\phi }_{i+1}^{T}\boldsymbol{\phi }_{i+1} \right ).
\end{align}
Using (\ref{equ_9})-(\ref{equ_14}), the Bayesian hypothesis test in (\ref{equ_8}) can be simplified to give the final activity rule for $w_{i+1}$ as 
\begin{equation}
\label{equ_15}
\mathrm{Activity}_{\mathrm{START}}\left ( w_{i+1} \right )\triangleq\mathbf{x}^{T}\boldsymbol{\phi}_{i+1}\boldsymbol{\phi}^{T}_{i+1}\mathbf{x}> \mathrm{Th}_{1,i+1},
\end{equation}
where $\mathrm{Th}_{1,i+1}$ is defined as 
\begin{equation}
\label{equ_16}
\mathrm{Th}_{1,i+1}\triangleq 2\sigma _{n}^{2}\left ( 1+\frac{\sigma_{n}^{2} }{\sigma_{\theta }^{2} } \right )\ln\left ( \frac{p_{00}}{p_{10}}\sqrt{\left ( 1+\frac{\sigma_{n}^{2} }{\sigma_{\theta }^{2} } \right )} \right ),
\end{equation}
 and $\mathbf{x}=\mathbf{y}-\mathbf{\Phi }\mathbf{w}-\boldsymbol{\phi }_{i}w_{i}-\boldsymbol{\phi }_{i+1}w_{i+1}$. It is seen that in the activity rule $\mathrm{Activity}_{\mathrm{START}}\left ( w_{i+1} \right )$ in (\ref{equ_15}) the correlation between the columns of matrix $\mathbf{\Phi}$ and measurement vector $\mathbf{x}$ decides between $\mathcal{H}_{01}$ and
$\mathcal{H}_{00}$.

\subsubsection{Searching The Termination of Active Blocks}
The detection of the end of an active block is performed by choosing one between the hypotheses $\mathcal{H}_{10}: s_{i}=1,s_{i+1}=0$ and $\mathcal{H}_{11}: s_{i}=1,s_{i+1}=1$, given the measurement vector $\mathbf{y}$. The Bayesian hypothesis test is given as
\begin{equation}
\label{equ_17}
\widehat{s}_{j}=
\begin{cases}
0 &  p\left ( \mathcal{H}_{10,j}\mid \mathbf{y} \right )>p\left ( \mathcal{H}_{11,j}\mid \mathbf{y} \right ), \\
1 &  Otherwise.
\end{cases}
\end{equation}
Similar to (\ref{equ_6}), we have $p\left ( \mathcal{H}_{10,j}\mid \mathbf{y} \right )=(1-p) \times p_{01} \times p\left ( \mathbf{y}\mid s_{i}s_{i+1}=10\right )$, where 
$\mathbf{y}=\sum_{j=1,j\neq i+1 }^{M}\boldsymbol{\phi}_{j}w_{j}+\mathbf{n}$. Likewise, $p\left ( \mathcal{H}_{11,j}\mid \mathbf{y} \right )=(1-p) \times p_{11} \times p\left ( \mathbf{y}\mid s_{i}s_{i+1}=11\right )$, where $p_{11}=p\left ( s_{i+1}=1 \mid s_{i}=1 \right )=1-p_{01}$ and $\mathbf{y}=\sum_{j=1}^{M}\boldsymbol{\phi}_{j}w_{j}+\mathbf{n}$. Therefore, we have the following inactivity rule for $w_{i+1}$
\begin{equation}
\label{equ_18}
p_{01} \times p\left ( \mathbf{y}\mid s_{i}s_{i+1}=10\right )>p_{11} \times p\left ( \mathbf{y}\mid s_{i}s_{i+1}=11\right ).
\end{equation}
Similar to (\ref{equ_9}), the likelihood function $p\left (\mathbf{y}\mid s_{i}s_{i+1} =10  \right )$ is calculated as
\begin{equation}
\label{equ_19}
\begin{split}
p\left (\mathbf{y}| s_{i}s_{i+1} =10  \right )&=\frac{\exp\left (-\frac{1}{2\sigma_{n}^{2} }\left \|\mathbf{y}-\sum_{j=1,j\neq i+1 }^{M}\boldsymbol{\phi}_{j} w_{j}  \right \|_{2}^{2}  \right )}{\sqrt{\left (2\pi\sigma_{n}^{2}   \right )^{N}}}.
\end{split}
\end{equation}
Also, given $s_{i}s_{i+1}=11$, $\mathbf{y}=\sum_{j=1,j\neq i+1}^{M}\boldsymbol{\phi}_{j}w_{j}+\boldsymbol{\phi}_{i+1}w_{i+1}+\mathbf{n}=\sum_{j=1,j\neq i+1}^{M}\boldsymbol{\phi}_{j}w_{j}+{\mathbf{n}}'$, where ${\mathbf{n}}'=\boldsymbol{\phi}_{i+1}w_{i+1}+\mathbf{n}$. Hence, the likelihood $p\left ( \mathbf{y}\mid s_{i}s_{i+1}=11\right )$ is a multivariate Gaussian with its covariance given by (\ref{equ_11})  and its mean by
\begin{equation}
\label{equ_20}
{\boldsymbol{\mu}}'_{\mathbf{y}}=\sum_{j=1,j\neq i+1}^{M}\boldsymbol{\phi}_{j}w_{j}.
\end{equation}
Also, the likelihood function $p\left ( \mathbf{y}\mid s_{i}s_{i+1}=11\right )$ can be evaluated as
\begin{equation}
\label{equ_21}
p\left ( \mathbf{y}\mid s_{i}s_{i+1}=11\right )=\frac{\exp\left(-\frac{1}{2}\left ( \mathbf{y}- {\boldsymbol{\mu}}'_{\mathbf{y}}\right )^{T}{\boldsymbol{\Sigma}^{-1}_{\mathbf{y}}}\left ( \mathbf{y}- {\boldsymbol{\mu}}'_{\mathbf{y}}\right )\right)}{\sqrt{\left ( 2\pi \right )^{N}\det(\boldsymbol{\Sigma}_{\mathbf{y}})}},
\end{equation}
where ${\boldsymbol{\Sigma}^{-1}_{\mathbf{y}}}$ and $\det\left ( {\boldsymbol{\Sigma}_{\mathbf{y}}} \right )$ are given in (\ref{equ_13}) and (\ref{equ_14}), respectively. 
Substituting (\ref{equ_19}) and (\ref{equ_21}) in (\ref{equ_18}) and using (\ref{equ_20}), the final inactivity rule for $w_{i+1}$ can be expressed as
\begin{equation}
\label{equ_22}
\mathrm{Inactivity}_{\mathrm{END}}\left ( w_{i+1} \right )\triangleq\mathbf{z}^{T}\boldsymbol{\phi}_{i+1}\boldsymbol{\phi}^{T}_{i+1}\mathbf{z}> \mathrm{Th}_{2,i+1},
\end{equation}
where $\mathrm{Th}_{2,i+1}$ is defined as 
\begin{equation}
\label{equ_23}
\mathrm{Th}_{2,i+1}\triangleq 2\sigma _{n}^{2}\left ( 1+\frac{\sigma_{n}^{2} }{\sigma_{\theta }^{2} } \right )\ln\left ( \frac{p_{01}}{p_{11}}\sqrt{\left ( 1+\frac{\sigma_{n}^{2} }{\sigma_{\theta }^{2} } \right )} \right ),
\end{equation}
 and $\mathbf{z}=\mathbf{y}-\mathbf{\Phi }\mathbf{w}-\boldsymbol{\phi }_{i+1}w_{i+1}$.

Also, the estimates of the unknown parameters $\sigma _{n}$, $\sigma _{\theta }$, $p$, $p_{10}$, and $p_{01}$ in (\ref{equ_16}) and (\ref{equ_23}) are given by the following
simple updates \cite{refe_BHT_5}
\begin{equation}
\label{equ_24}
\hat{\sigma }_{n}=\frac{\left \| \mathbf{y}-\mathbf{\Phi }\hat{\mathbf{w}} \right \|_{2}}{\sqrt{N}},
\hat{\sigma }_{\theta}=\sqrt{\frac{N\mathbb{E}(y_j^2)}{M(1-\hat{p})}},
\hat{p}= \frac{\left \| \mathbf{s} \right \|_{0}}{M},
\end{equation}
\begin{equation}
\label{equ_25}
\hat{p}_{10}= \frac{\sum_{i=1}^{M-1}s_{i+1}\left ( 1-s_{i} \right )}{\sum_{i=1}^{M-1}\left ( 1-s_{i} \right )},
\hat{p}_{01}= \frac{\sum_{i=1}^{M-1}s_{i}\left ( 1-s_{i+1} \right )}{\sum_{i=1}^{M-1}s_{i}},
\end{equation}
where $\mathbb{E}\left ( \cdot \right )$ represents the expectation of a random variable.

\subsection{Amplitude Estimation Using Linear MMSE}\label{sec_amp}
Given the detection and recovery information of the binary support vector $\mathbf{s}$ by BHT, we complete the estimation of the original unknown signal $\mathbf{w}$ by estimating 
the amplitude samples of the $\boldsymbol{\theta}$ vector.

Based on the detected vector $\mathbf{s}$, denoted by $\hat{\mathbf{s}}$, we obtain the linear MMSE estimate (\cite{refe_BHT_6}, p. 364) of $\boldsymbol{\theta}$ (denoted by 
$\hat{\boldsymbol{\theta}}$) which is given as
\begin{equation}
\label{equ_26}
\hat{\boldsymbol{\theta}}=\sigma_{\theta}^2\hat{\mathbf{S}}\boldsymbol{\Phi}^T\left ( \sigma^{2}_{n}\mathbf{I}_{N}+\sigma^{2}_{\theta}\boldsymbol{\Phi} 
\hat{\mathbf{S}} \boldsymbol{\Phi}^{T}\right )^{-1}\mathbf{y}
\end{equation}
where $\hat{\mathbf{S}}=\diag(\hat{\mathbf{s}})$.

\textbf{Algorithm~\ref{fig: alg}}  provides a pseudo-code implementation of our proposed Block-BHTA that gives all steps in the algorithm including BHT support detection and amplitude
estimation.
\begin{algorithm}
\caption{The overall Block-BHTA estimation.}
\label{fig: alg}
{\fontsize{7.5}{10}\selectfont
\begin{algorithmic}[1]
\REQUIRE $\mathbf{y}$, $\mathbf{\Phi}$, $k_{max}$, and $\epsilon$
Initialize: $\text{Choose}\hspace{1mm} p^{\left ( 0 \right )} \in [0.5,1]$, $\sigma ^{\left ( 0 \right )}_{\theta}= \sqrt{\frac{N\mathbb{E}(y_j^2)}{M(1-\hat{p})}}$, $\sigma ^{\left ( 0 \right )}_{n}= \sigma ^{\left ( 0 \right )}_{\theta}/5$, $\mathbf{w}^{(0)}=\mathbf{\Phi}^T(\mathbf{\Phi}\mathbf{\Phi}^T)^{-1}\mathbf{y}$.
set $\mathrm{difference}=1$, $k=0$.
\WHILE{$\left ( \mathrm{difference}> \epsilon \quad \mathrm{and} \quad k< k_{\mathrm{max}} \right )$}
\STATE \textbf{BHT-detection:}
\FOR{$i=0,\cdots ,M-1$}
\IF{$\mathrm{Activity}_{\mathrm{START}}\left ( w_{i+1} \right )>\mathrm{Th}_{1,i+1}$ in (\ref{equ_15})}
\STATE \quad set \quad$s_{i}=1$,
\ELSIF{$\mathrm{Inactivity}_{\mathrm{END}}\left ( w_{i+1} \right )>\mathrm{Th}_{2,i+1}$ in (\ref{equ_22})}
\STATE\quad set \quad$s_{i}=0$,
\ENDIF
\ENDFOR \\
\STATE \textbf{LMMSE estimation:}\quad$\mathbf{S}^{\left ( k \right )}=\diag\left ( \mathbf{s}^{\left ( k \right )} \right )$,
\STATE$\boldsymbol{\theta}^{\left ( k \right )}= \sigma_{\theta}^2\hat{\mathbf{S}}\boldsymbol{\Phi}^T\left ( \sigma^{2}_{n}\mathbf{I}_{N}+\sigma^{2}_{\theta}\boldsymbol{\Phi} 
\hat{\mathbf{S}} \boldsymbol{\Phi}^{T}\right )^{-1}\mathbf{y}$,
\STATE \textbf{Parameter Estimation:}\quad
 using (\ref{equ_24}) and (\ref{equ_25})
\STATE $\mathbf{w}^{\left ( k \right )}=\mathbf{s}^{\left ( k \right )}\odot\boldsymbol{\theta }^{\left ( k \right )}$.
\STATE \text{Compute the difference}\,\,$\triangleq\frac{\left \| \mathbf{w}^{\left ( k+1 \right )}- \mathbf{w}^{\left ( k \right )}\right \|_{2}}{\left \| \mathbf{w}^{\left ( k+1 \right )} \right \|_{2}}$, $k\leftarrow k+1$
\ENDWHILE
\ENSURE $\widehat{\mathbf{w}}=\mathbf{w}^{\left ( k \right )}$
\end{algorithmic}
}
\end{algorithm}
\section{Simulation Results}\label{sec_sim}
This section presents the experimental results to demonstrate the performance of the Block-BHTA. Two experimental results are presented in this section. First, we compare the performance of the proposed Block-BHTA with that of BPA \cite{refe_BHT_3} versus SNR. Second, we evaluate the performance of Block-BHTA versus number of nonzero blocks and compare the performance with
some block-sparse signal reconstruction algorithms.

All the experiments are conducted for 400 independent simulation runs.
In each simulation run, the elements of the matrix $\boldsymbol{\Phi}$ are chosen from a uniform distribution in [-1,1] with columns normalized to unit \ltwo-norm. The
Block-sparse sources $\mathbf{w_{gen}}$ are synthetically generated using BGHMM in (\ref{equ_4}) which is based on Markov chain process. Unless otherwise stated,
in all experiments $p=0.9$, $p_{01}=0.09$ and $\sigma_{\theta}=1$ which are the parameters of BGHMM. The measurement vector
$\mathbf{y}$ is constructed by $\mathbf{y}=\mathbf{\Phi}\mathbf{w_{gen}}+\mathbf{n}$, where $\mathbf{n}$ is zero-mean AWGN with a variance tuned to a specified value of SNR
which is defined as
\begin{equation}
\label{equ_SNR}
\mathrm{SNR\left ( dB \right )}\triangleq20\log_{10}\left ( \left \| \mathbf{\Phi}\mathbf{w_{gen}} \right \|_{2}/\left \| \mathbf{n} \right \|_{2} \right ).
\end{equation}
We use the Normalized Mean Square Error (NMSE (dB)) as a performance metric, defined by $\mathrm{NMSE (dB)}\triangleq10\log_{10}(\left \| \widehat{\mathbf{w}}-\mathbf{w_{gen}} \right \|^{2}_{2}/\left \|\mathbf{w_{gen}} \right \|^{2}_{2})$,
where $\widehat{\mathbf{w}}$ is the estimate of the true signal $\mathbf{w_{gen}}$.
\begin{figure}[tb]
\centering
\includegraphics[width=5.9cm]{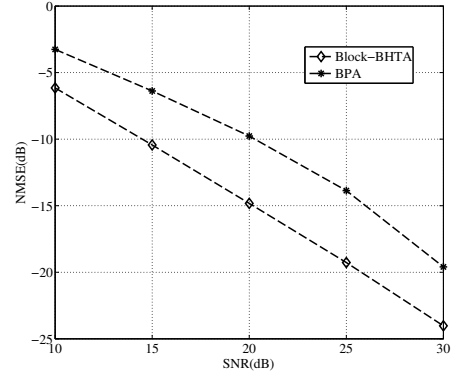}
\caption{\footnotesize NMSE (dB) versus SNR for Block-BHTA and BPA. The results are averaged over 400 trials.}
\label{fig1}
\end{figure}
We compare the Block-BHTA and BPA at different noise levels. In this experiment $N=192$ and $M=512$. We add the Gaussian white noise so that SNR, defined in (\ref{equ_SNR}), 
varies between 10 dB and 30 dB for each generated signal.

Figure \ref{fig1} shows the NMSE (dB) versus SNR for both Block-BHTA and BPA. It is seen that Block-BHTA
exhibit significant performance gain (almost 5 dB) over BPA.

In the second experiment, we examine the influence of the block size and the number of blocks on the estimation performance of the Block-BHTA where the block partition is unknown. Towards that end, we set up a simulation to compare the Block-BHTA with some recently developed algorithms for 
block sparse signal reconstruction, such as the block sparse Bayesian learning algorithm (BSBL) \cite{refe_11}, the expanded  block sparse Bayesian learning algorithm (EBSBL) \cite{refe_11}, the cluster-structured MCMC algorithm (CluSS-MCMC) \cite{refe_9}, and the pattern-coupled sparse Bayesian learning  algorithm (PC-SBL) \cite{refe_BHT_7}.
The size of matrix $\boldsymbol{\Phi}$ is $256 \times 512$, $\mathrm{SNR}=15 \mathrm{dB}$, and  $\sigma_{\theta}=1$.
\begin{figure}[tb]
\centering
\includegraphics[width=5.9cm]{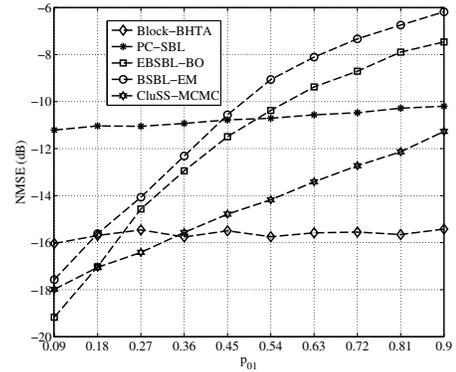}
\caption{\footnotesize NMSE versus $p_{01}$ for Block-BHTA and other algorithms. The results are averaged over 400 trials.}
\label{fig2}
\end{figure}
Recall from Section \ref{sec_sigmodel} that the block size and the number of blocks of $\mathbf{w}$ are proportional to $1/p_{01}$. That is, when $p_{01}$ is small $\mathbf{w}$
comprises small number of blocks with big sizes and vice versa. Hence, we vary the value of $p_{01}$ between $0.09$ and $0.9$ to obtain the NMSE (dB) for various algorithms.
The results of NMSE (dB) versus $p_{01}$ is shown in Fig. \ref{fig2}. As seen from the figure, for $p_{01}\geq0.36$ the Block-BHTA
outperforms all other algorithms.

\section{Conclusion}\label{sec_conc}
This letter has presented a novel Block-BHTA to recover the block-sparse signals whose structure of block sparsity is completely unknown. The proposed Block-BHTA uses a Bayesian 
hypothesis testing to detect and recover the support of the block sparse signal. For amplitude recovery, Block-BHTA utilizes a linear MMSE to estimate the nonzero amplitudes of the 
detected supports. Simulation results demonstrate that Block-BHTA outperforms the BPA by almost 5 dB performance gain. The Block-BHTA also outperforms many state-of-the-art algorithms when the block-sparse signal comprises a large number of blocks with short lengths.

\ifCLASSOPTIONcaptionsoff
  \newpage
\fi



%

\bibliographystyle{IEEEtran}
\bibliography{B-BHTA}



%








\end{document}